\documentclass{jbi}
\usepackage{lipsum} 
\usepackage{booktabs} 
\usepackage{multirow} 
\usepackage{xspace}
\usepackage{smile}
\usepackage{amsfonts}
\usepackage{hyperref}
\usepackage{url}

\usepackage[superscript,biblabel,compress]{cite}

\usepackage{xcolor,colortbl}
\usepackage{adjustbox}
\usepackage{subfig}
\usepackage{tablefootnote}

\usepackage[edges]{forest}
\usepackage{tabularray}
\usepackage{tikz}
\usetikzlibrary{chains,shapes.geometric}

\newif\ifsubmit
\submitfalse
\definecolor{darkpink}{rgb}{0.91, 0.33, 0.5}

\ifsubmit

\else

\fi

\begin{document}

\title{A Review on Knowledge Graphs for Healthcare: \\Resources, Applications, and Promises}

\author{Hejie Cui, PhD$^1$, Jiaying Lu, PhD$^2$, Shiyu Wang, PhD$^1$, Ran Xu, MS$^1$, Wenjing Ma, PhD$^3$, Yue Yu, MS$^4$, Shaojun Yu, MS$^1$, Xuan Kan, PhD$^1$, Chen Ling, MS$^1$, Tianfan Fu, PhD$^5$, Liao Zhao, PhD$^1$, Joyce C. Ho, PhD$^1$, Fei Wang, PhD$^6$, Carl Yang, PhD$^1$}

\institutes{
    $^1$ Department of Computer Science, Emory University, Atlanta, GA, USA  \\
    $^2$ Nell Hodgson Woodruff School of Nursing’s Center for Data Science, Emory University, Atlanta, GA, USA \\
    $^3$ Department of Biostatistics, University of Michigan, Ann Arbor, MI, USA \\
    $^4$ School of Computational Science and Engineering, Georgia Institute of Technology, Atlanta, GA, USA \\
    $^5$ Department of Computational Science, Rensselaer Polytechnic Institute, Troy, NY, USA \\
    $^6$ Department of Population Health Sciences, Weill Cornell Medicine, Cornell University, Ithaca, NY, USA
}

\maketitle

\begingroup
\onehalfspacing
\begin{tabular}{@{}p{0.23\textwidth} p{0.72\textwidth}@{}}
    \textbf{Authors:} 
    & Hejie Cui, PhD$^{1*}$ \\
    & Jiaying Lu, PhD$^{1*}$ \\
    & Ran Xu, MS$^{1*}$ \\
    & Shiyu Wang, PhD$^1$ \\
    & Wenjing Ma, PhD$^2$ \\
    & Yue Yu, PhD$^3$ \\
    & Shaojun Yu, MS$^1$ \\
    & Xuan Kan, PhD$^1$ \\
    & Chen Ling, PhD$^1$ \\
    & Liang Zhao, PhD$^1$ \\
    & Zhaohui S. Qin, PhD$^1$\\
    & Joyce C. Ho, PhD$^1$ \\
    & Tianfan Fu, PhD$^4$ \\
    & Jing Ma, PhD$^5$\\
    & Mengdi Huai, PhD$^6$\\
    & Fei Wang, PhD$^7$ \\
    & Carl Yang, PhD$^1$ \\
\end{tabular}
\endgroup

$^*$Equal contributions.

\begingroup
\onehalfspacing
\begin{tabular}{@{}p{0.23\textwidth} p{0.72\textwidth}@{}}
    \textbf{Affiliation of the authors:} & $^1$ Department of Computer Science, Emory University, Atlanta, GA, USA \\
    & $^2$ Department of Biostatistics, University of Michigan, Ann Arbor, MI, USA  \\
    & $^3$ School of Computational Science and Engineering, Georgia Institute of Technology, Atlanta, GA, USA \\
    & $^4$ Department of Computational Science, Rensselaer Polytechnic Institute, Troy, NY, USA \\
    & $^5$ Department of Computer and Data Sciences, Case Western Reserve University, Cleveland, OH, USA \\
    & $^6$ Department of Computer Science, Iowa State University,  Ames, IA, USA \\
    & $^7$ Department of Population Health Sciences, Weill Cornell Medicine, Cornell University, Ithaca, NY, USA \\
\end{tabular}
\endgroup

\begingroup
\onehalfspacing
\begin{tabular}{@{}p{0.23\textwidth} p{0.72\textwidth}@{}}
    \textbf{Correspondence:} & Carl Yang, PhD \\
    & 400 Dowman Drive, Suite W401 \\
    & Atlanta, GA, 30322  \\
    & Phone: +1 (217) 417-5987 \\
    & Email: j.carlyang@emory.edu \\
\end{tabular}
\endgroup


\newpage

\section*{Abstract}

\textbf{Objective:} This comprehensive review aims to provide an overview of the current state of Healthcare Knowledge Graphs (HKGs), including their construction, utilization models, and applications across various healthcare and biomedical research domains.

\textbf{Methods:} We thoroughly analyzed existing literature on HKGs, covering their construction methodologies, utilization techniques, and applications in basic science research, pharmaceutical research and development, clinical decision support, and public health. The review encompasses both model-free and model-based utilization approaches and the integration of HKGs with large language models (LLMs).

\textbf{Results:} We searched Google Scholar for relevant papers on HKGs and classified them into the following topics: HKG construction, HKG utilization, and their downstream applications in various domains. We also discussed their special challenges and the promise for future work.

\textbf{Discussion:} The review highlights the potential of HKGs to significantly impact biomedical research and clinical practice by integrating vast amounts of biomedical knowledge from multiple domains. The synergy between HKGs and LLMs offers promising opportunities for constructing more comprehensive knowledge graphs and improving the accuracy of healthcare applications.

\textbf{Conclusions:} HKGs have emerged as a powerful tool for structuring medical knowledge, with broad applications across biomedical research, clinical decision-making, and public health. 
This survey serves as a roadmap for future research and development in the field of HKGs, highlighting the potential of combining knowledge graphs with advanced machine learning models for healthcare transformation.

\newpage
\section{Introduction}
\label{sec:introduction}

A knowledge graph (KG) is a data structure that captures the relationships between different entities and their attributes~\cite{ji2021survey,nicholson2020constructing}. KG models and integrates data from various sources, including structured and unstructured data, and has been studied to support a wide range of applications such as search engines~\cite{wang2019knowledge}, recommendation systems~\cite{wang2019kgat,zhou2020interactive}, and question answering~\cite{lin2019kagnet,yasunaga2021qa,yan2021learning,kan2021zero}. Healthcare Knowledge Graph (HKG) facilitates an interpretable representation of medical concepts, e.g., drugs and disease, as well as the relations among those medical concepts. This data structure enables the connection of contexts and enhances clinical research and decision-making~\cite{santos2022knowledge, primekg}.

On the data side, HKG is usually built on complex medical systems such as electronic health records, medical literature, clinical guidelines, and patient-generated data~\cite{bouayad2017patient,rajkomar2018scalable}. However, these data resources are often heterogeneous and distributed, which makes it challenging to integrate and analyze them effectively~\cite{mehta2018concurrence}. This data heterogeneity can also lead to incomplete or inconsistent data representations, limiting their usefulness for downstream healthcare tasks~\cite{dash2019big}. Additionally, the current use of domain-specific KGs may result in limited coverage and granularity of the knowledge captured across different levels. This hinders identifying correlations and relationships between medical concepts from multiple domains. These challenges highlight the need for continued research on HKGs to realize their full potential. 

On the modeling side, the construction of HKGs can be done either from scratch or by integrating existing dataset resources. Many crucial steps, such as entity and relation extraction, can be done with NLP tools and algorithms.
Recently, there have been significant advancements in general domain knowledge extraction, thanks to pre-trained large language models (LLMs) such as BERT~\cite{devlin2019bert}, GPT Series~\cite{brown2020language}, and others. 
These models revolutionize the field and make it possible to integrate heterogeneous medical data from various sources effectively. The use of pre-trained models has also led to the development of more accurate and comprehensive medical ontologies and taxonomies~\cite{zhang2021taxonomy,yu2020steam,wang2021enquire,zeng2021enhancing, xu2020building}. This allows for the evaluation of generated contents from LLMs and reduces LLM hallucination.

A comprehensive HKG has the potential to contribute to health research across various levels~\cite{gyrard2018personalized,santos2022knowledge,li2020real}. At the micro-scientific level, HKGs can help researchers identify new phenotypic and genotypic correlations and understand the underlying mechanisms of disease~\cite{hassani2017knowledge}, leading to more targeted and effective treatments~\cite{seneviratne2021personal,primekg}. At the clinical care level, HKGs can be used to develop clinical decision support systems that provide clinicians with relevant information, improving clinical workflows and patient outcomes~\cite{eberhardt2012clinical,castaneda2015clinical}. Therefore, a thorough review of existing literature on HKGs becomes an essential roadmap and invaluable resource to drive transformative advancements in the field.

\textbf{Statement of Significance:} While prior reviews have focused on KGs tailored for specific healthcare tasks such as drug discovery~\cite{bonner2022review}, electronic health record (EHR) predictions~\cite{murali2023towards}, and ontologies~\cite{silva2022ontologies}, they often lack a holistic perspective across diverse healthcare domains. General surveys on KG construction~\cite{ji2021survey} provide foundational insights but do not delve into the unique challenges and requirements of the healthcare context. Furthermore, some studies~\cite{abu2023healthcare} concentrate solely on KG construction without exploring the critical dimension of knowledge utilization in healthcare applications. This work bridges these gaps by offering a comprehensive review that not only synthesizes knowledge graph construction techniques but also emphasizes their practical applications across various healthcare domains, providing a roadmap for future innovation.

\begin{figure*}[h!]
\centering
\includegraphics[width=\textwidth]{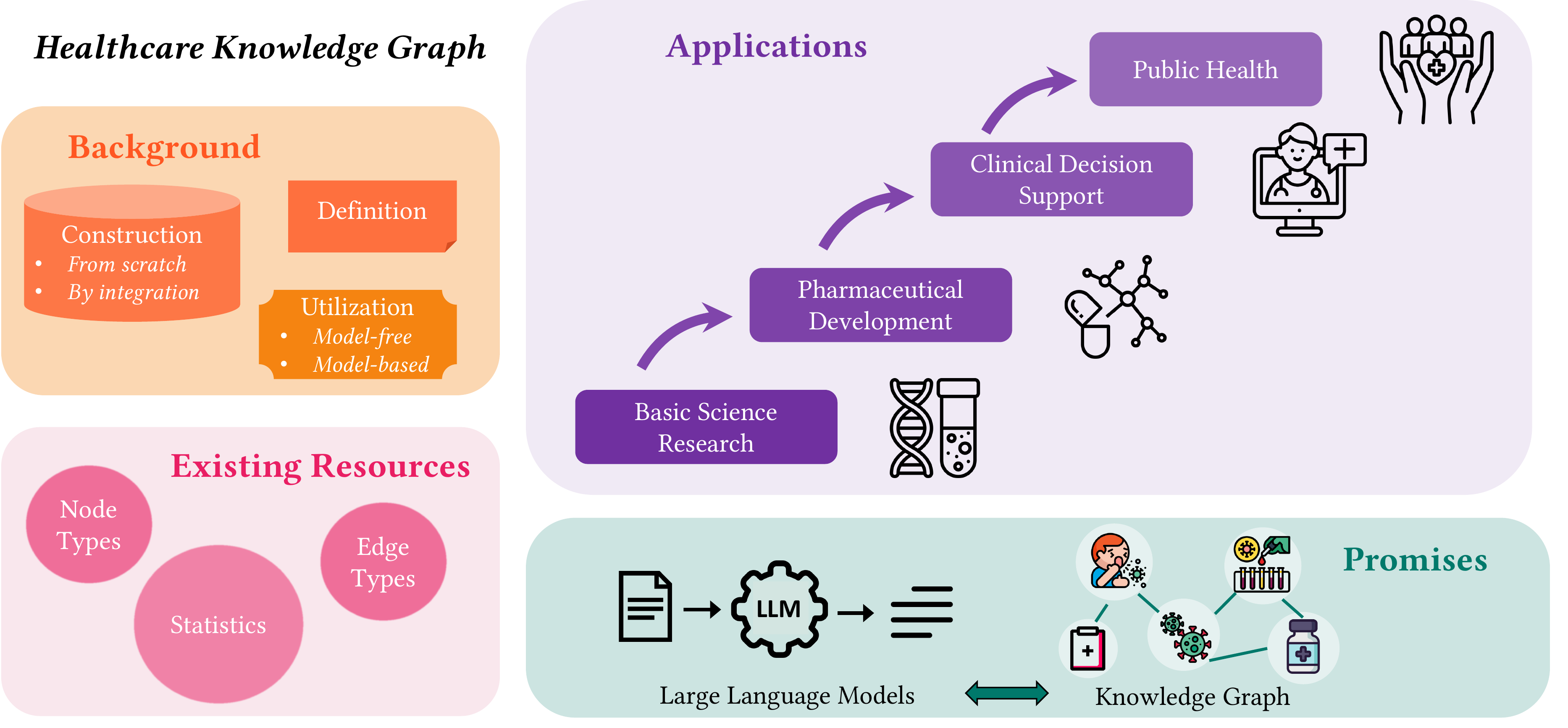}
\caption{The content overview of this review on healthcare knowledge graphs.}
\label{fig:fig_taxo}
\end{figure*}

\textbf{Literature Selection.
} We conducted a systematic literature review following the PRISMA (Preferred Reporting Items for Systematic Reviews and Meta-Analyses) guidelines to identify and include studies relevant to the development and application of HKGs. Searches were conducted across multiple academic databases, including Google Scholar, PubMed, IEEE Xplore, ACM Digital Library, ACL Anthology, and Web of Science. Search terms were crafted to capture the scope of HKG research comprehensively. These terms included combinations of the following keywords using Boolean operators (AND, OR, NOT):

Healthcare Knowledge Graphs: ``healthcare knowledge graph" ``HKG"  ``Ontology" ``medical knowledge graph". \\
Applications: ``clinical decision support" ``drug repurposing" ``pharmaceutical development" 
``clinical trial matching" ``bioinformatics". \\
Construction Techniques: ``ontology integration" ``entity resolution" ``relation extraction" ``knowledge graph embeddings". \\
Integration with AI: ``large language models" ``machine learning" ``natural language processing". \\
Reference lists of key articles were manually searched to identify additional relevant studies. The search timeframe spanned from 2010 to 2024 with 362 articles in total. 
We then removed duplicate articles based on their titles and authors. After deduplication, we examined the remaining articles' titles and abstracts for topic relevance.
Finally, for each section, we conducted a detailed examination of the methodology described in each paper to ensure relevance, excluding those deemed irrelevant. After the filtering process, a total of 175 papers were retained.

\begin{figure*}[h!]
\centering
\includegraphics[width=0.9\textwidth]{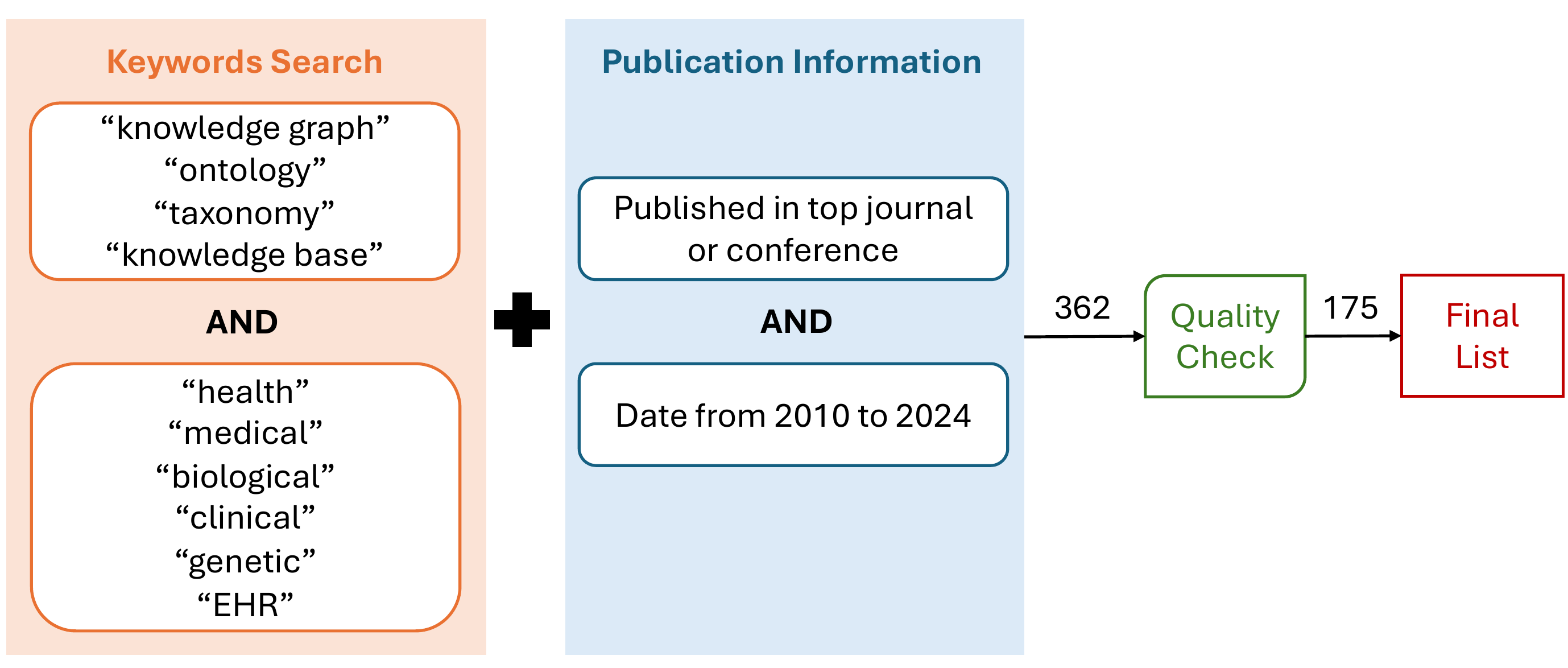}
\caption{The PRISMA diagram illustrating the literature search process.}
\vspace{-10pt}
\label{fig:fig_taxo}
\end{figure*}


\section{Backgrounds}
\label{sec:background}


\noindent\textbf{Healthcare Knowledge Graphs (HKGs).} 
We focus on HKG, which is a structured representation of knowledge that captures entities (e.g., organizations, diseases, medications) and the relationships between them. KGs typically employ graph-based structures, where entities are represented as nodes and relationships as edges, and may use attributes or properties for additional contexts. These graphs can be enriched with contextual information and metadata, enabling complex relationships and data interconnectivity. Besides, we also include ontologies and knowledge bases, which are commonly used in constructing HKGs. An ontology is a formal model of a domain's concepts, properties, and relationships, typically using a hierarchical or taxonomical structure, with an emphasis on semantic relation, and interoperability of different healthcare concepts. In contrast, KGs integrate data from heterogeneous sources, scale dynamically, and are well-suited for data-driven insights and probabilistic reasoning. By leveraging both ontologies and KGs, HKGs can support robust knowledge discovery and reasoning for healthcare applications. Furthermore, combining these resources with LLMs can enhance decision support by grounding LLM outputs in structured, contextualized knowledge and reduce the risk of hallucination. By covering these categories of terminology, we provide a comprehensive overview of the different types of resources available for organizing and representing medical knowledge in a structured and semantically rich manner.

\begin{figure*}[!htbp]
\centering
\forestset{
  L1/.style={draw=black, text width=3cm},
  L2/.style={draw=black, text width=4cm},
  L3/.style={draw=black, text width=5cm},
  L4/.style={draw=black, text width=5cm},
}
\begin{forest}
for tree={draw,rounded corners,grow'=0,text width=2cm, text centered,edge+={->}},
forked edges,
[Background, fill=orange!30, L1
[HKG Definition, fill=orange!20, L2]
[HKG Construction, fill=orange!20, L2
    [Construct from Scratch, fill=orange!10, L3]
    [Construct by Integration, fill=orange!10, L3]
]
[HKG Utilization, fill=orange!20, L2
    [Model-free Utilization, fill=orange!10, L3]
    [Model-based Utilization, fill=orange!10, L3]
]
]
\end{forest}
\caption{Detailed taxonomy of the background section of healthcare knowledge graphs.}
\label{fig:taxonomy-background}
\end{figure*}
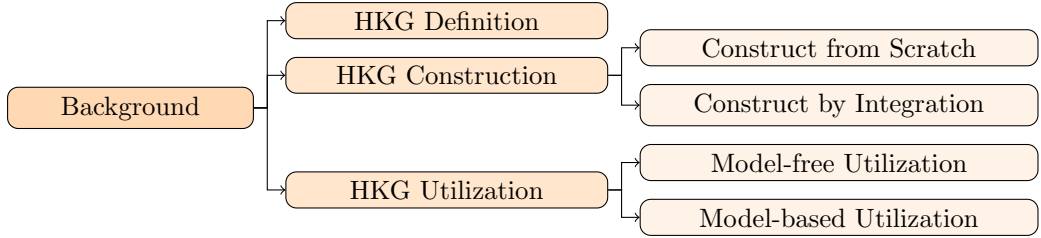

\subsection{HKG Construction}

\begin{figure*}[htbp!]
\centering
\begin{tikzpicture}[start chain,node distance=5mm,
                    every join/.style={->}]
  \node [draw,on chain,join] {Develop schema};
  \node [draw,on chain,join] {Collect data};
  \node [draw,on chain,join] {Extract knowledge triples};
  \node [draw,on chain=going below,
         join] {Normalize entities and relations};
  \node [draw,on chain=going left,
         join] {Infer missing links};
  \node [draw,on chain=going left,
         join] {Update and validate};
\end{tikzpicture}
\caption{The pipeline of constructing HKGs from scratch.}
\label{fig:HKG_convention_construction}
\end{figure*}
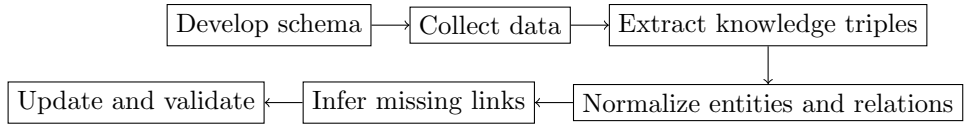

\textbf{Constructing HKGs from Scratch.} 
A multi-step pipeline, as in Figure~\ref{fig:HKG_convention_construction}, is used to construct HKGs from scratch.
\begin{enumerate}
    \item The first step is to identify the scope and objectives. Researchers typically develop a schema~\cite{guarino2009ontology,blagec2022curated} or use existing schemas~\cite{guarino2009ontology,ashburner2000GO,schriml2012DO,bard2005CO} to clearly specify the domain, ensuring consistent and aligned knowledge. Unlike the general domain KG, utilizing schemas is a common practice in HKG construction.
    \item Secondly, researchers gather data from various sources, including medical literature, clinical trials, and patient-generated data. It is essential to ensure the quality and consistency of the data and to remove identifiable information for patient privacy.
    \item The third step involves transforming the data into a structured format. This includes identifying medical entities and creating relationships between them via specialized biomedical NLP tools~\cite{song2021BNER,xing2020biorel,hahn2020medical}.
    \item Entities and relationships are then mapped to selected ontologies using thesauruses~\cite{bodenreider2004UMLS} or terminologies~\cite{donnelly2006snomed,hirsch2016icd} to ensure the KG is compatible with other healthcare systems and supports data integration.
    \item Once the initial KG is built, the next step is to fill in missing links between entities using graph databases~\cite{wang2020txt2sql} or link prediction models~\cite{bordes2013transE,lu22:hakeGCN}.
    \item The final step is to continuously update and validate the KG to ensure accuracy and relevance. This step involves incorporating new data and knowledge, refining the schema, and evaluating the quality of the KG.
\end{enumerate}

\textbf{Constructing HKGs by Integration.}
Considering significant efforts have been paid to construct and curate HKGs from scratch, it is promising to integrate these data resources to avoid repetitive work. HKG integration (also called HKG fusion) refers to the processing of merging two or more HKGs into a single, more comprehensive graph~\cite{himmelstein2017systematic,ibkh,youn2022knowledge}. The integration process is challenging because different HKGs may use different terminologies, schemas, or data formats. To address these challenges, researchers have developed various techniques and algorithms for KG fusion, including ontology matching~\cite{faria2014towards,he2022machine}, schema alignment~\cite{suchanek2011paris,maaroufi2014formalizing}, entity resolution~\cite{bachman2018famplex,hu2021integrated}, and conflicts resolution~\cite{ma2023tecre}. These methods aim to identify and reconcile the differences between KGs.

\textbf{Techniques for HKG Constructions.}
Traditionally, each step of HKG construction involves one specially designated model. For instance, Hidden Markov Models and Recurrent Neural Networks are widely used for healthcare named entity recognition, relation extraction tasks, while Translational Models and Graph Neural Networks are used for HKG completion and conflict resolution tasks.
Recently, LLMs have shown great utility to serve as a uniform tool for constructing KGs~\cite{ye2022GenerativeKGCons}. Several key steps of constructing KGs, such as named entity recognition~\cite{liang2020bond,chen2023one,huang2022copner,liu2022qaner}, relation extraction~\cite{zhuang-etal-2022-resel,lu2022UIE,yang2022large}, entity linking~\cite{de2022autogreEL,mrini2022multiEL,cho2022unsupervisedEL}, and KG completion~\cite{saxena2022sequenceKGC,xie2022discriminationKGC,shen2022paltKGC}, have been successfully tackled by these large foundation models. 
Early explorations of construction HKG with large foundation models show that healthcare entity normalization~\cite{zhang2021graphprompt,agrawal2022large}, healthcare entity recognition~\cite{fries2021ontology,hu2023zero}, healthcare entity linking~\cite{zhu2022BEL}, and healthcare knowledge fusion~\cite{lu23HiPrompt} can also be performed, without extensive training on healthcare corpus.
On the other hand, researchers start to construct KGs under the open-world assumption~\cite{shi2018open,das2020probabilistic,niu2021open,OERL,lu22:hakeGCN}, thus getting rid of the dependency on pre-defined schemas and exhaustive entity\&relation normalization. 

\begin{table}
\scriptsize
\caption{Resources of existing HKGs. }
\label{tab:resource}
\centering
\begin{tabular}{cp{0.18\columnwidth}p{0.21\columnwidth}p{0.18\columnwidth}p{0.17\columnwidth}}
\toprule
\bf Name & \bf Node Types & \bf Edge Types & \bf Statistic & \bf Application  \\
\midrule
HetioNet~\cite{himmelstein2015heterogeneous} &  11 (e.g., drug, disease) & 24 (e.g., drug-disease) & \#N: 47.0 K, \#E: 2.3 M  &  Medicinal Chemistry  \\
DrKG~\cite{drkg2020} & 13 (e.g., disease, gene) & 107 (e.g., disease-gene) & \#N: 97 K, \#E: 5.8 M & Medicinal Chemistry  \\
PrimeKG~\cite{primekg} & 10 (e.g., phenotypes) & 30 (e.g., disease-phenotype)  & \#N: 129.4 K, \#E: 8.1 M  & Medicinal Chemistry    \\
Gene Ontology\tablefootnote{\url{http://geneontology.org/}}~\cite{ashburner2000GO} & 3 (e.g., biological process) & 4 (e.g., partOf) & \#N: 43 K, \#E: 7544.6K &  Bioinformatics  \\
KEGG\tablefootnote{\url{https://www.genome.jp/kegg/}}~\cite{kanehisa2000kegg} & 16 (e.g., pathway) & 4 (e.g., partOf) & \#N: 48.5 M, \#E: unknown &   Bioinformatics \\
STRING\tablefootnote{\url{https://string-db.org/}}~\cite{szklarczyk2023string} & 1 (e.g., protein) & 4 (e.g., interactions) & \#N: 67.6 M, \#E: 20 B &  Bioinformatics \\
Cell Ontology\tablefootnote{\url{https://www.ebi.ac.uk/ols4/ontologies/cl}}~\cite{diehl2016cell} & 1 (i.e., cell type) & 2 (e.g, subClassOf) & \#N: 2.7 K, \#E: 15.9 K  & Bioinformatics  \\
GEFA~\cite{ranjan2022generating} & 510 (e.g., kinases) & 2 (e.g., drug-drug) & \#N: 0.5 K, \#E: 30.1 K & Drug Development   \\
Reaction~\cite{li2022prediction} & 2 (e.g., reactant \& normal) & 19 (e.g., reaction paths) & \#N: 2192.7 K, \#E: 932.2 K & Drug Development   \\
ASICS~\cite{jeong2022intelligent} & 2 (e.g., reactant \& product) & 1 (e.g., reactions) & \#N: 1674.9 K, \#E: 923.8 K & Drug Development   \\
Hetionet~\cite{jeong2022intelligent} & 11 (e.g., biological process) & 24 (e.g., disease–associates–gene) & \#N: 47.0 K, \#E: 2250.2 K & Drug Development   \\
LBD-COVID~\cite{zhang2021drugcov} & 1 (i.e., concept) & 1 (i.e., SemMedDB relation) & \#N: 131.4 K, \#E: 1016.1 K & Drug Development   \\
GP-KG~\cite{gao2022kg} & 7 (e.g., drug) & 9 (e.g., disease–gene) & \#N: 61.1 K, \#E: 1246.7 K & Drug Development   \\
DRKF~\cite{zhang2021drug} & 4 (e.g., drug) & 43 (e.g., drug-disease) & \#N: 12.5 K, \#E: 165.9 K & Drug Development   \\
DDKG~\cite{ghorbanali2023drugrep} & 2 (i.e., drug \& disease) & 1 (e.g., drug-disease) & \#N: 551, \#E: 2.7 K & Drug Development   \\
Disease Ontology\tablefootnote{\url{https://disease-ontology.org/}}~\cite{schriml2012DO} & 1 (i.e., disease) & 2 (e.g., subClassOf) & \#N: 11.2 K, \#E: 8.8 K & Clinical Decision Support  \\
DrugBank~\cite{wishart2018drugbank} & 4 (e.g., drug, pathway) & 4 (e.g., drug-target) & \#N: 7.4 K, \#E: 366.0 K & Clinical Decision Support     \\
KnowLife~\cite{ernst2014knowlife} & 6 (e.g., genes) & 14 (e.g., gene-diseases)  & \#N: 2.9 M, \#E: 11.4 M  & Clinical Decision Support \\
PharmKG~\cite{zheng2021pharmkg} & 3 (e.g., diseases)& 3 (e.g., chemical-diseases) &  \#N: 7601,	\#E: 500958 & Clinical Decision Support \\
ROBOKOP\tablefootnote{\url{https://robokop.renci.org/}}~\cite{bizon2019robokop} & 54 (e.g., genes, drugs) & 1064 (e.g., biolink, CHEBI) & \#N: 8.6M , \#E: 130.4 M & Clinical Decision Support \\
iBKH\tablefootnote{\url{https://github.com/wcm-wanglab/iBKH}}~\cite{ibkh} & 11 (e.g., anatomy, disease) & 18 (e.g., anatomy-gene) & \#N: 2.4 M, \#E: 48.2 M & Clinical Decision Support   \\
\bottomrule
\end{tabular}
\label{tab:stats}
\end{table}

\subsection{HKG Utilization}

\textbf{Model-free Utilization.}
Various query languages can be used for KGs, such as SPARQL, Cypher, and GraphQL~\cite{wang2020txt2sql}. These query languages allow users to query HKGs using a standardized syntax, thus enabling users to retrieve, manipulate, and analyze data in a structured and consistent way. 
For instance, automatic healthcare question answering can be tackled by Natural Language Question-to-Query
(NLQ2Query) approach~\cite{pmlr-v174-kim22a}, where natural language questions are first translated into executable graph queries and then answered by the query responses.
HKGs can also be utilized as an up-to-date and trustworthy augmentation to LLMs for many applications. Some pioneering studies~\cite{guu2020retrieval,xu2023weakly,shi2023replug} show that retrieved knowledge triples can improve the reliability of LLMs in various knowledge-intensive tasks. 
Moreover, KGs can be a useful tool for fact-checking~\cite{tchechmedjiev2019claimskg,vedula2021face,mayank2022deapFaked} as they provide a structured representation of information that can be used to quickly and efficiently verify the accuracy of claims. Researchers have explored the utility of HKGs in identifying ingredient substitutions of food~\cite{shirai2021identifying}, COVID-19 fact-checking~\cite{mengoni2022COVDfact}, etc.

\textbf{Model-based Utilization.}
Utilizing HKGs in complex reasoning tasks often involves utilizing machine learning models. HKG embeddings~\cite{yu2021sumgnn,su2022attention} have shown great potential to tackle these tasks. In particular, HKG embedding models are a class of machine learning models that aim to learn low-dimensional vector representations of the entities and relations in a KG. After obtaining HGK embeddings, they can be plugged into any kind of deep neural network and further fine-tuned toward downstream objectives.
On the other hand, Symbolic logic models offer an interpretable approach to KG reasoning by mining logical rules from existing knowledge through techniques such as inductive logic programming~\cite{muggleton1992inductive}, association rule mining~\cite{galarraga2013amie}, or Markov logic networks~\cite{kok2005learning}. These minded rules are used to infer new facts, make logical deductions and answer complex queries. Recently, researchers start to explore combining logical rules into KG embedding to further improve the generalization and performance of HKG reasoning~\cite{alshahrani2017neuro,zhu2022neural}.


\section{Applications}
\label{sec:application}

\begin{figure*}[!htbp]
\centering
\forestset{
  L1/.style={draw=black, text width=2.5cm},
  L2/.style={draw=black, text width=4.2cm},
  L3/.style={draw=black, text width=4.5cm},
  L4/.style={draw=black, text width=4.5cm},
}
\begin{forest}
for tree={draw,rounded corners,grow'=0,text width=2cm, text centered,edge+={->}},
forked edges,
[Application, fill=blue!30, L1
[Basic Science Research, fill=blue!20, L2
    [Medical Chemistry, fill=blue!10, L3]
    [Bioinformatics, fill=blue!10, L3]
]
[Pharmaceutical Development, fill=blue!20, L2
    [Drug Development, fill=blue!10, L3]
    [Clinical Trial, fill=blue!10, L3]
]
[Clinical Decision Support, fill=blue!20, L2
    [Intermediate Steps to Advance Prediction Models, fill=blue!10, L3]
    [Evidence Generation for Risk Prediction, fill=blue!10, L3]
]
[Public Health, fill=blue!20, L2
    [Epidemiology, fill=blue!10, L3]
    [Environmental Health, fill=blue!10, L3]
    [Policy and Management, fill=blue!10, L3]
    [Social and Behavioral, fill=blue!10, L3]
]
]
\end{forest}
\caption{Detailed taxonomy of the application section of healthcare knowledge graphs.}
\vspace{-3mm}
\label{fig:taxonomy-application}
\end{figure*}
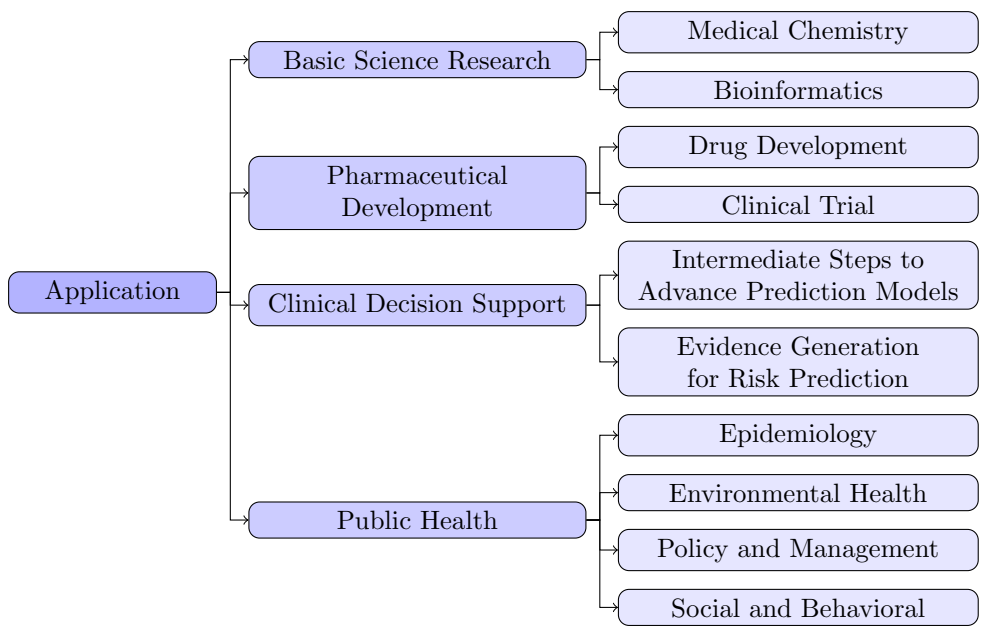

\subsection{Basic Science Research}
Biologists have built various ontologies (gene, cell, disease) and networks (gene regulatory, protein interaction) over the past decade, which are also KGs. They are commonly used in medicinal chemistry and bioinformatic research.

\subsubsection{Medicinal Chemistry}

\noindent \textbf{Drug-drug interactions (DDIs)} refer to changes in the actions, or side effects, of drugs when they are taken at the same time or successively~\cite{giacomini2007good}. In general, DDIs are a significant contributor to life-threatening adverse events~\cite{su2022attention}, and their identification is one of the key tasks in public health and drug development. The existence of diverse datasets on drug-drug interactions (DDIs) and biomedical KGs has enabled the development of machine-learning models that can accurately predict DDIs. Yu et al.~\cite{yu2021sumgnn} develop SumGNN, a model that efficiently extract relevant subgraphs from a KG to generate reasoning paths within the subgraph, resulting in significantly improved predictions of multi-typed DDIs. Su et al.~\cite{su2022attention} propose DDKG, an attention-based KG representation learning framework that involves an encoder-decoder layer to learn the initial embeddings of drug nodes from their attributes in the KG. Karim et al.~\cite{karim2019drug} compare various techniques for generating KG embeddings with different settings and conclude that a combined convolutional neural network and LSTM network yields the highest accuracy when predicting drug-drug interactions (DDIs). Dai et al.~\cite{dai2021drug} propose a new KG embedding framework by introducing adversarial autoencoders based on Wasserstein distances and Gumbel-Softmax relaxation for DDI tasks. Lin et al.~\cite{lin2020kgnn} develop KGNN that resolves the DDI prediction by capturing drugs and their potential neighborhoods by mining associated relations in KG.

\noindent \textbf{Drug-target interactions (DTIs)} are as important as DDIs~\cite{chen2016drug}. Machine learning models can leverage KGs constructed from various types of interactions, such as drug-drug, drug-disease, protein-disease, and protein-protein interactions, to aid in predicting DTIs. For instance, Li et al.~\cite{li2023ga} utilize the KG transfer probability matrix to redefine the drug-drug and target-target similarity matrix, thus constructing the final graph adjacent matrix to learn node representations by utilizing dual Wasserstein Generative Adversarial Network. 
Zhang et al.~\cite{zhang2021discovering} propose a new hybrid method for DTI prediction by first constructing DTI-related KGs and then employing graph representation learning model to obtain feature vectors of the KG. Wang et al.~\cite{wang2022kg} construct a KG of 29,607 positive drug-target pairs by DistMult embedding strategy, and propose a Conv-Conv module to extract features of drug-target pairs. Ye et al.~\cite{ye2021unified} learn a low-dimensional representation for various entities in the KG, and then integrate the multimodal information via neural factorization machine.

\subsubsection{Bioinformatics Research}

\noindent \textbf{Multi-Omics Analysis} has become increasingly important for understanding complex biological systems. With the advancement of high-throughput technologies, 
more KG applications based on multi-omics data integration have emerged, aiming to provide new research methods to uncover the complex relationships between different omics layers and reveal biological systems' underlying mechanisms.

KGs have been used to identify disease-associated mutations, genes, proteins, and metabolites by integrating multi-omics data with existing biological knowledge. 
This approach has led to the discovery of novel biomarkers and therapeutic targets for various diseases~\cite{zhang2021ddn2}. 
Quan et al. built a comprehensive multi-relational KG called AIMedGraph, providing an interpretation of the impact of genetic variants on disease or treatment~\cite{quanAIMedGraphComprehensiveMultirelational2023}. They curated detailed information about diseases, drugs, genetic variants, and the impact of genetic variations on disease development and drug treatment from multiple data resources. 
GenomicsKG is a KG to analyze and visualize multi-omics data. GenomicsKG can be used to improve drug development based on clinical genomics correlations and personalized drug customization in the extended version based on interactive relationships. It also provides multi-dimensional visualization, linked functional KGs, and reporting for clinical genomics.

\noindent \textbf{Single-Cell Analysis} focuses on cells as the fundamental and essential units of living organisms. With high-throughput sequencing technologies advancing to measure genomic profiles in a single-cell resolution, cell functions (inside cells) and cell-cell interactions (between cells) are revealed~\cite{linnarsson2016single}. 
Gene regulatory mechanisms, which control gene expression and affect processes like cell differentiation and disease progression, are key to understanding these functions.
Traditional gene knockdown experiments are time-consuming and limited, but single-cell sequencing provides genome-wide data, including gene expression, transcription factor binding, DNA methylation, and epigenetic modifications. This allows researchers to uncover gene regulatory networks (GRNs) that enhance our understanding of biological processes. Databases like GRNdb and GenomicKB integrate sequencing data and annotations to provide insights into gene regulation across tissues~\cite{fang2021grndb,feng2023genomickb}, which will improve as more data becomes available.

\subsection{Pharmaceutical Research Development}

\subsubsection{Drug Development}

Drug development is the process of identifying novel chemical compounds that can effectively treat or alleviate human diseases. Before the drug can be designated as a final product for clinical use, several critical steps need to be undertaken from the initial target identification, chemical synthesis and clinical trials. The whole process typically spans over a decade and involves expenses of approximately one billion dollars~\cite{adams2006estimating}, yet it is characterized by a low success rate for clinical approval~\cite{dimasi2015cost}.

\noindent \textbf{Drug Design} is an area where KGs are commonly used, particularly for generating novel molecules that are promising drug candidates for various diseases~\cite{ranjan2022generating, li2022prediction}. Ranjan et al.~\cite{ranjan2022generating} utilize Gated Graph Neural Network (GGNN) to generate novel molecules that target the coronavirus (i.e., SARS-CoV-2)~\cite{hasoksuz2020coronaviruses} and integrate KGs into their approach to reduce the search space. Specifically, KGs were leveraged to discard non-binding molecules before inputting them into the Early Fusion model, thus optimizing the efficiency of the drug design process. 
In addition to employing deep learning for direct structure design, KGs are also utilized in the analysis of chemical synthesis. 
Quantitative estimation of molecular synthetic accessibility is critical in prioritizing the molecules generated from generative models. 
For instance, Li et al.~\cite{li2022prediction} utilize reaction KGs to construct classification models for compound synthetic accessibility. By leveraging KGs that capture information about reactions, including reaction types, substrates, and reaction conditions, they can train machine learning models that could predict the synthetic accessibility of compounds. Jeong et al.~\cite{jeong2022intelligent} introduce an intelligent system that integrates generative exploration and exploitation of reaction knowledge base to support synthetic path design.

\noindent \textbf{Drug Repurposing} has often been expedited by the utilization of KGs~\cite{zhu2020knowledge, maclean2021knowledge, himmelstein2017systematic, zhang2021drugcov, gao2022kg, xu2019network, zhang2021drug, ghorbanali2023drugrep}. Many applications on drug re-purposing that utilize KGs are primarily focused on link prediction tasks~\cite{maclean2021knowledge}. To re-purpose promising drug candidates for new indications, many methods employ predictive models that focus on predicting drug-treats-disease relationships within pharmacological KGs. 
Xu et al.~\cite{xu2019network} develop a multi-path random walk model on a network that incorporates gene-phenotype associations, protein-protein interactions, and phenotypic similarities for training and prediction purposes. Zhang et al.~\cite{zhang2021drugcov} introduce an integrative and literature-based discovery model for identifying potential drug candidates from COVID-19-focused research literature, including PubMed and other relevant sources. Gao et al.~\cite{gao2022kg} construct a KG by integrating multiple genotypic and phenotypic databases. They then learn low-dimensional representations of the KG and utilize these representations to infer new drug-disease interactions, providing insights into potential drug repurposing opportunities. Zhang and Che~\cite{zhang2021drug} introduce a model for drug re-purposing in Parkinson's disease that leverages a local medical knowledge base incorporating accurate knowledge along with medical literature containing novel information. Ghorbanali et al.~\cite{ghorbanali2023drugrep} present the DrugRep-KG method, which utilizes a KG embedding approach for representing drugs and diseases in the process of drug repurposing.

\subsubsection{Clinical Trial}
The major goal of clinical trials is to assess the safety and effectiveness of drug molecules on human bodies. 
A novel drug molecule needs to pass three phases of clinical trials before it is approved by the Food and Drug Administration (FDA) and enters the drug market. 
The whole process is prohibitively time-consuming and expensive, costing 7-11 years and two billion dollars on average~\cite{martin2017much}. 

\noindent \textbf{Clinical Trial Optimization} targets identifying eligible patients for clinical trials based on their medical history and health conditions~\cite{rivera2020guidelines,he2020clinical}. 
Recently, with massive electronic health records (EHR) data and trial eligibility criteria (EC),  data-driven methods have been studied to automatically assign appropriate patients for clinical trials~\cite{yuan2019criteria2query,tseo2020information,liu2021evaluating}. However, it is often hard to fully capture and represent the complex knowledge present in unstructured ECs and EHR data, as ECs may only provide general disease concepts. In contrast,  patient EHR data contain more specific medical codes to represent patient conditions.
To better capture the interactions among different medical concepts from EHR records and ECs,  Gao et al.~\cite{gao2020compose} enhance patient records with hierarchical taxonomies to align medical concepts of varying granularity between EHR codes and ECs. Besides, Fu et al.~\cite{fu2022hint} leverage additional knowledge-embedding modules along with drug pharmacokinetic and historical trial data to improve the patient trial optimization process, and Wang et al.~\cite{wang2023spot} leverage the KGs to learn static trial embedding and further designed meta-learning module to generalize well over the imbalanced clinical trial distribution.

\subsection{Clinical Decision Support}
\label{sec:clinical_support}
Electronic Health Record (EHR) contains essential patient information such as disease diagnoses, prescribed medications, and test results. 
However, the sparsity of EHR data largely restricts the ability of deep learning approaches.
To overcome this drawback, KGs have been applied to incorporate prior medical knowledge for these deep learning models to better support the downstream prediction tasks.

\subsubsection{Intermediate Steps to Advance Prediction Models}
\textbf{ICD Coding} aims to extract diagnosis and procedure codes from clinical notes which are often raw texts~\cite{mullenbach2018explainable,zhang2020bert,vu2021label,dong2022automated}. 
It is often challenging, as the size of the candidate target codes can be large and the distribution of the codes is often long-tailed~\cite{pmlr-v149-kim21a}.
To overcome this, Xie et al.~\cite{xie2019ehr} and Cao et al.~\cite{cao2020hypercore} 
propose to leverage KGs as \emph{distant supervision}~\cite{min2013distant,liang2020bond}, and inject the label information via structured \emph{KG propagation} by leveraging  graph convolution networks~\cite{kipf2016semi} to learn the correlations among medical codes.
Besides, Lu et al.~\cite{lu2020multi} propose to leverage KGs and the co-occurrence graph among clinical nodes simultaneously with a knowledge aggregation module to boost the ICD coding performance. Ren et al.~\cite{pmlr-v182-ren22a} design a learning curriculum based on the hierarchical structure of the code to address the highly imbalanced label distribution issue and balance between frequent and rare labels. Overall, injecting additional knowledge with graph neural networks
offers a way to mitigate the imbalanced label distribution issue and thus better. \\
\textbf{Entity and Relation Extraction} helps convert the rich unstructured or semi-structured data in health records text into structured data that can be more easily processed, understood, and utilized by clinicians and algorithms.
Specifically, \emph{entity extraction} aims to identify entity mentions from clinical-free texts. 
There are two key steps for entity extraction, i.e., named entity recognition (NER) and disambiguation (NED). By leveraging additional KGs, 
Yuan et al.~\cite{yuan2023exploring} inject additional knowledge from the KGs for entity linking and proposed post-pruning and Thresholding to improve the efficiency and remove the effect of unlinkable entity mention.
Fries et al.~\cite{fries2021ontology} leverage clinical ontologies to provide \emph{weak supervision} sources to create additional training data for clinical entity disambiguation.
Besides, \emph{relation extraction} aims to identify and classify relationships between entities in unstructured text, which facilitates understanding complex biological processes, drug interactions, and disease mechanisms. 
To incorporate the external KG, several works~\cite{fei2021enriching,roy2021incorporating} proposed additional post-training steps to align the language models with biomedical knowledge.
Hong et al.~\cite{hong2021clinical} construct embeddings for a wide range of codified concepts from EHRs to identify relevant features related to a disease of interest, and Lin et al.~\cite{lin2022multimodal} design a co-training scheme to jointly learn from text and KGs for extracting and classifying disease-disease relations. In summary, fusing KGs with language models can flexibly accommodate missing data types and bring additional performance gains, especially for those rare entities and relations.

\subsubsection{Evidence Generation for Risk Prediction Models}

\textbf{Disease Prediction} aims to predict the potential diseases of a given patient with his past clinical records. To assist the diagnosis with additional knowledge, GRAM~\cite{choi2017gram} and KAME~\cite{ma2018kame} utilize a medical ontology~\cite{dubberke2006icd} where the leaf nodes are the medical codes found in EHR data, and their ancestors are more general categories. 
By incorporating information from medical ontologies into deep learning models via neural attention, these approaches learn better embeddings for different medical concepts to alleviate the data scarcity bottleneck.
\cite{yin2019domain,zhang2019knowrisk} further consider the domain-specific KG KnowLife~\cite{ernst2015knowlife} to enrich the embeddings of medical entities with their neighbors on the KG. 
These approaches mainly directly update the embeddings of different concepts to improve the feature learning, but may be at the risk of ignoring the high-level order information from the KG. 
To tackle this drawback, Ye et al.~\cite{ye2021medpath} explicitly exploit \emph{paths} in KG from the observed symptoms to the target disease to model the personalized information for diverse patients. 
Xu et al.~\cite{xu2021predictive} design a self-supervised learning approach to pre-train
a graph attention network for learning the embedding of medical concepts and completing the KG simultaneously.
These approaches better harness the structure information, and often lead to better performance than the pure embedding-based knowledge integration techniques.

\noindent\textbf{Treatment Recommendation} aims to recommend personalized medications to patients based on their individual health conditions, which can help physicians select the most effective medications for their patients, and improve treatment outcomes~\cite{zhang2017leap,bhoi2021personalizing,shang2019gamenet}. To effectively exploit external knowledge, Shang et al.~\cite{shang2019pre} use drug ontologies to design additional pretraining loss and directly improve the representation of drugs, and several studies~\cite{wu2022leveraging,tan20224sdrug} attempt to extract the additional drug interaction graphs to model the negative side effects of specific drug pairs and reduce the possibility of recommending negative drug-drug interaction combinations. 
Besides, 
Wu et al.~\cite{wu2022knowledge} leveraged ontologies to improve the drug representations, and facilitates drug recommendation under a more challenging few-shot setting.

\subsection{Public Health}
Public Health research can significantly benefit from HKGs. KGs can help organize, structure, and formalize extensive information from diverse and heterogeneous sources. This allows researchers to analyze data, reason about factors, and make decisions on a larger scale.

\noindent\textbf{Epidemiology}. The field of epidemiology has seen an increased use of KGs to analyze and understand the spread of diseases. A study by Gao et al.~\cite{gao2023knowledge} analyzes the research and development trends of wastewater-based epidemiology (WBE) using KGs constructed from nearly 900 papers. Domingo-Fern{\'a}ndez et al.~\cite{domingo2021covid} create the COVID-19 KG, a comprehensive cause-and-effect network constructed from the scientific literature on the coronavirus. Additionally, Turki et al.~\cite{turki2022using} use KGs to assess and validate the portion of Wikidata related to COVID-19 epidemiology using an automatable task set. 
Pressat Laffouilh{\`e}re et al.~\cite{pressat2022ontological} develop OntoBioStat, a domain ontology related to covariate selection and bias in biostatistics, which can help interpret significant statistical associations between variables.

\noindent\textbf{Environmental Health}. Fecho et al.~\cite{fecho2021biomedical} develop ROBOKOP, a biomedical KG-based system, to validate associations between workplace chemical exposures and immune-mediated diseases. Wolffe et al.~\cite{wolffe2020survey} propose using KGs in systematic evidence mapping in environmental health. This approach overcomes the limitations of rigid data tables by offering a more suitable model for handling the highly connected and complex nature of environmental health data.

\noindent\textbf{Health Policy and Management}. Wu et al.~\cite{wu2021construct} have analyzed the COVID-19 epidemic situation using a KG of patient activity. This method enables in-depth study of the transmission process, analysis of key nodes, and tracing of activity tracks. Meanwhile, Yu et al.~\cite{yu2022improving} develop a chronic management system, which combines KGs and big data to optimize the management of chronic diseases in children. This system enhances treatment and resource utilization while conforming to the requirements of the Chronic Care Model.

\noindent\textbf{Social and Behavioral Health}. Cao et al.~\cite{cao2020building} build a high-level suicide-oriented KG combined with deep neural networks for detecting suicidal ideation on social media platforms. 
Also, Liu et al.~\cite{liu2020knowledge} conduct a bibliometric analysis of driver behavior research. Additionally, Wang et al.~\cite{wang2022interpretable} create an analysis framework for interpreting causal associations in emotional logic. They introduce a KG into appraisal theories, improving human emotional inference.

\begin{table}
\scriptsize
\caption{Main challenges in different types of problems. }
\label{tab:challenge}
\centering
\begin{tabular}{cp{0.67\columnwidth}}
\toprule
\bf Types of Problems & \bf Main Challenges \\
\midrule
Basic Science Research &  - Limited integration of KGs with LLMs for structural and functional predictions of molecules and genes~\cite{soman2024biomedical}. \newline
- Difficulty in representing high-dimensional data (e.g., protein structures) in KGs~\cite{daza2023bioblp}. \newline
- Lack of domain-specific fine-tuning of LLMs for bioinformatics tasks~\cite{keloth2024advancing}.  \\
\midrule
Pharmaceutical Development & - Incomplete or inconsistent drug-related knowledge representation in KGs~\cite{jiang2024geometric}. \newline
- Challenges in combining structured KG data with unstructured LLM outputs (e.g., literature mining) for assisting drug discovery~\cite{ghorbanali2023drugrep}. \\
\midrule
Clinical Decision Support & 
- Difficulty in ensuring explainability and reliability of predictions~\cite{gaur2021semantics}. \newline
- Lack of harmonization between real-time clinical data and static KGs for dynamic decision-making~\cite{abu2023healthcare}.    \\
\midrule
Public Health & - Limited ability to model causal relationships and complex interactions in public health domains using KGs~\cite{jaimini2022causalkg}. \newline
- Difficulty in integrating population-level data with KGs for real-time surveillance.~\cite{yang2021constructing}  \\
\bottomrule
\end{tabular}
\label{tab:challenge}
\end{table}

\section{Challenge, Promise, and Outlook}
\label{sec:promise}
\textbf{Current Challenges for Adopting KGs to Biomedical Research.} Injecting KGs into clinical problems faces significant challenges across domains. In basic science research, the integration of KGs and LLMs for molecular predictions is limited by difficulties in representing high-dimensional data like protein structures and a lack of domain-specific fine-tuning~\cite{soman2024biomedical, daza2023bioblp, keloth2024advancing}. In pharmaceutical development, inconsistent drug-related knowledge representation and challenges in combining KGs with unstructured LLM outputs hinder drug discovery~\cite{jiang2024geometric, ghorbanali2023drugrep}. Clinical decision support struggles with ensuring explainability, reliability, and harmonizing real-time clinical data with static KGs for adaptive decision-making~\cite{gaur2021semantics, abu2023healthcare}. Public health faces limitations in modeling causal relationships and integrating population-level data with KGs for real-time surveillance due to scalability and standardization issues~\cite{jaimini2022causalkg, yang2021constructing}. Addressing these barriers is vital for leveraging KGs to advance clinical and public health outcomes.

\textbf{Future Directions} The potential impact of comprehensive and fine-grained HKGs on biomedical research and clinical practice is significant. By integrating vast amounts of biomedical knowledge from multiple domains, HKGs can facilitate the discovery of new disease mechanisms and the identification of novel drug targets. They also help to enable personalized medicine by identifying patient subgroups with shared disease mechanisms. 
The recent success of LLMs such as ChatGPT offers promising opportunities in capturing such semantics from the biomedical context~\cite{agrawal2022large,singhal2022large,nath2022new,moor2023foundation}, enabling the construction of unprecedentedly comprehensive HKGs. In turn, HKGs also help improve LLMs by providing accurate and contextualized knowledge to regularize the generated content. This is particularly useful in evaluating LLMs in biomedical applications and addressing the problem of hallucination in critical areas. 
For developers of novel biomedical informatics methods, this symbiotic relationship between HKGs and LLMs presents both opportunities and challenges. One key area for innovation is the design of frameworks that allow real-time updating of HKGs with the latest clinical and research data, enabling dynamic decision-making. Developers must also address the computational and representational challenges of integrating high-dimensional data, such as genomic sequences and protein structures, into HKGs while ensuring semantic consistency. Moreover, constructing HKGs that incorporate causal reasoning capabilities could significantly enhance their utility in understanding complex disease mechanisms and predicting therapeutic outcomes. 
Another promising avenue lies in leveraging fine-tuned LLMs to extract domain-specific insights from unstructured biomedical literature and seamlessly integrate them into HKGs. This process, however, requires rigorous validation pipelines to ensure the accuracy and relevance of the extracted information. Developers should also explore strategies for aligning LLM outputs with ontologies and existing knowledge schemas to maximize interoperability and reuse.  
Finally, as HKGs are increasingly utilized to support clinical decision-making, developers must prioritize explainability and fairness in their design. This includes developing visualization tools and interactive interfaces that allow clinicians and researchers to interpret the outputs of HKG-powered systems effectively. Similarly, integrating multilingual capabilities and context-aware reasoning into HKGs will be essential for addressing global healthcare challenges and ensuring equitable access to cutting-edge biomedical insights.

\section{Conclusion}
\label{sec:conclu}
HKGs have emerged as a promising approach for capturing and organizing medical knowledge in a structured and interpretable way. This comprehensive review paper provides an overview of the current state of HKGs, including their construction, modeling, and applications in healthcare. Furthermore, the paper discusses potential future developments of HKGs.
In conclusion, HKGs have played a significant role in advancing health research. With the advent of LLMs, there are even more opportunities to combine HKGs and LLMs to reduce the generation of false or unreliable content. We hope that our comprehensive review of this field offers a helpful perspective for future reference. 

\section*{Acknowledgements}
This research was partially supported by the National Science Foundation under Award Number 2319449 and Award Number 2312502, as well as the National Institute Of Diabetes And Digestive And Kidney Diseases of the National Institutes of Health under Award Number K25DK135913. Any opinions, findings, and conclusions or recommendations expressed herein are those of the authors and do not necessarily represent the views, either expressed or implied, of the National Science Foundation, National Institutes of Health, or the U.S. government.
The authors wish to thank the editors and reviewers for their valuable efforts and suggestions.

\newpage

\makeatletter
\renewcommand{\@biblabel}[1]{\hfill #1.}
\makeatother

\bibliographystyle{vancouver}
\bibliography{jbi}

\end{document}